\documentclass{article}
\usepackage{spconf,amsmath,graphicx}
\usepackage{hyperref}
\usepackage{multirow}

\title{LEARNING TO WRITE WITH COHERENCE FROM NEGATIVE EXAMPLES}

\threeauthors
 {Seonil Son\sthanks{The first author performed this work while at Seoul National University.}}
	{Language AI Lab\\
	NCSOFT\\
	Gyeonggi-Do, S. Korea\\
	deftson@ncsoft.com}
 {Jaeseo Lim$^\ddagger$, Youwon Jang$^\dagger$, Jaeyoung Lee$^\dagger$}
	{$^\ddagger$Interdisciplinary Program in Cognitive Science, \\
	$^\dagger$School of Computer Science and Engineering\\
	Seoul National University, Seoul, S. Korea\\
	\{jaeseolim, sharifa, jerry96\}@snu.ac.kr}
 {Byoung-Tak Zhang$^{\dagger \star}$\thanks{This work was partly supported by the Korean government (2015-0-00310-SW.StarLab (25 \%), 2017-0-01772-VTT (25 \%), 2018-0-00622-RMI (25 \%), 2019-0-01371-BabyMind (25 \%)).}}
	{
	$^\star$Artificial Intelligence Institute\\
	Seoul National University\\
	Seoul, S. Korea\\
	btzhang@snu.ac.kr}
\begin{document}
%\ninept

\maketitle
\begin{abstract}
\textit{Coherence} is one of the critical factors that determine the quality of writing. 
We propose writing relevance (WR) training method for neural encoder-decoder natural language generation (NLG) models which improves \textit{coherence} of the continuation by leveraging negative examples. 
WR loss regresses the vector representation of the context and generated sentence toward positive continuation by contrasting it with the negatives. 
We compare our approach with Unlikelihood (UL) training in a text continuation task on commonsense natural language inference (NLI) corpora to show which method better models the \textit{coherence} by avoiding unlikely continuations. 
The preference of our approach in human evaluation shows the efficacy of our method in improving \textit{coherence}.
\end{abstract}
\section{Introduction}
\label{sec:intro}

\label{intro}
An open ending leaves the readers lots of possibilities to imagine their own conclusion to the story.
Endings by each writer may vary, but most of the endings will show \textit{coherence} with the preceding context, and therefore end up forming a story.

While the encoder-decoder, or Seq2Seq framework~\cite{seq2seq} is designed to model the conditional likelihood of the decoded sentences given a context vector, it is not enough for modeling \textit{coherent} semantics of the continuation.
To this end, we introduce Writing Relevance (WR) training which models \textit{coherence}, “a fit of the text to its context”~\cite{coherence_cohesion_writingqual}.
WR training loss regresses the vector representation of the context and generated sentences toward the positive continuation example while separating it further from negative examples.

We experiment with text continuation on commonsense NLI corpora, namely the HellaSWAG~\cite{hellaswag} and Story Cloze Test~\cite{roc} datasets.
The model is asked to generate the continuation given the context sentences while the original datasets require to choose among the provided continuations. 
\textit{Coherence} of continuations is the key due to the large number of possible cohesive (but incoherent) continuations and arising uncertainty the corpora allows.
Both corpora contain human-written negative examples which are \textit{cohesive} but diverge from information relevant to the context and thus not \textit{coherent}.\footnote{Cohesion is not coherence~\cite{cohesion_v_coherence}. Cohesive ties (e.g., use of pronouns, parallelism) may work as minimal conditions for \textit{coherence}, but do not warrant delivering the topic consistently.} 
As each corpus originates from different text sources\footnote{the HellaSWAG corpus is collected from WikiHow, and ActivityNet Captions~\cite{activitynetcaptions}; the Story Cloze Test is a collection of human-written 5-sentence long stories.}, it helps to examine the extensible use of our training scheme to different domains. \\[0.5ex]

    \begin{figure}[t] % h t b 
    \includegraphics[width=\columnwidth]{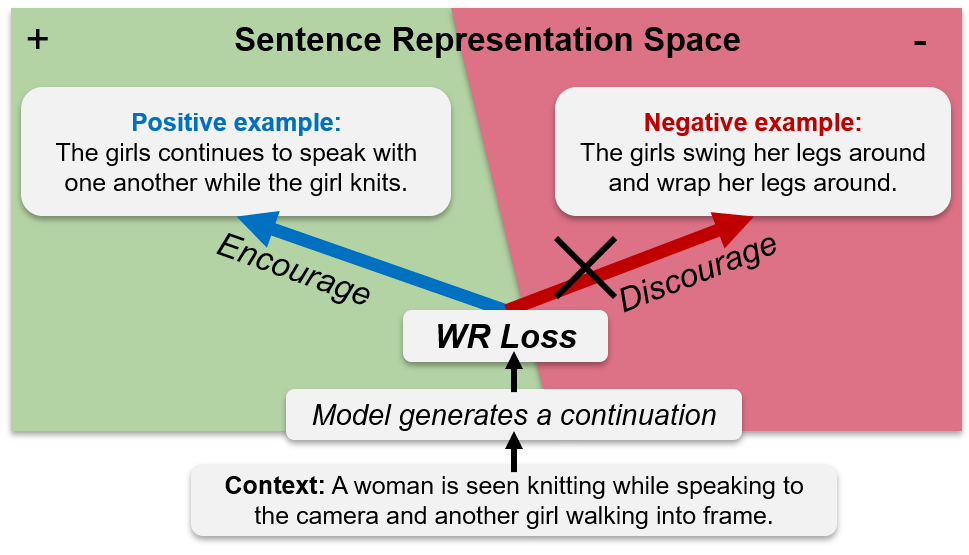}
    \caption{WR loss biases encoder-decoder NLG model to generate sentences closer to the positive than negative example in the representation space.}
    \label{fig:concept}
    \end{figure}

\noindent In summary, our contributions are two-fold:
\begin{itemize}
    \item We propose a WR training method that models \textit{coherence} of the continuation by leveraging negative examples.
    \item We demonstrate the potential of our WR training scheme to be applied to various domains by experimenting on two distinct commonsense NLI corpora.
\end{itemize}

\begin{figure*}[h!] \centering  \includegraphics[width=\linewidth ]{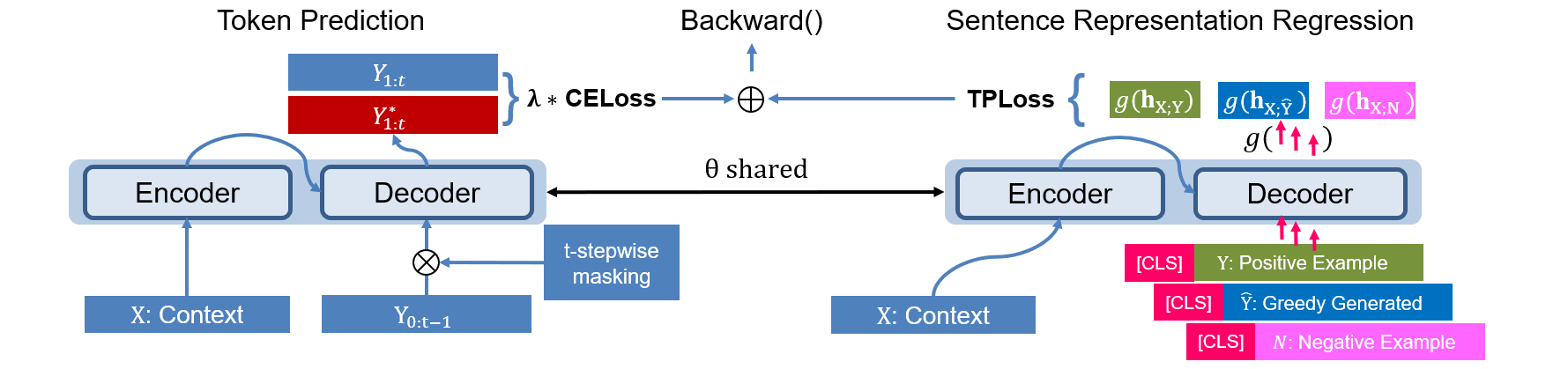}
\caption{Loss computation during the writing relevance (WR) training. Cross-entropy (CELoss) for token prediction is computed as in the pre-training, triplet loss (TPLoss) for contrasting negatives is added to the total loss.}
\label{fig:backprop}
\end{figure*}

\section{Learning Framework}

 This section covers the two-step procedure for WR training.
 First, we pre-train an encoder-decoder model to generate grammatically correct sentences on a American Literature Short Story (ALSS) corpus that we have collected.\footnote{Details of the collection and pre-processing can be found in \url{https://github.com/sonsus/american_literature}} 
 Then WR training takes place to fine-tune the model to bias its generation toward \textit{coherent} sentences while avoiding out-of-place sentences (i.e., negative examples).
 We describe these two steps of training for encoder-decoder model in Section \ref{pre-training} and \ref{wrtraining}, including details of the decoding procedure and sentence representation.
 
\subsection{Pre-training Encoder-Decoder Model}
\label{pre-training}
 Here, we train the encoder-decoder NLG model for maximum likelihood estimation (MLE) of the consecutive tokens given a context as: %in Equation \ref{eq:clm}
 
\begin{equation}
\label{eq:clm}
            P(Y|X) = \prod_{i}^{T}p(y_i|Y_{0:i-1}, X) 
\end{equation}

\noindent where $Y$ is a consecutive sentence that continues the context $X$. $X$ and $Y$ both consist of a series of tokens; \{$x_0$, $x_1$, ..., $x_{T'}$\}, \{$y_0$, $y_1$, ..., $y_T$\}. 
As we intend for the model to learn proper grammar for \textit{cohesion} but not to induce \textit{coherence}, we split the ALSS corpus into pairs of successive sentences. 

For pre-training to be effective for latter text continuation, we initially chose Toronto Boot Corpus~\cite{torontobook} as it is considered in-domain data for commonsense NLI corpora we are targeting.
However, to circumvent the copyright issue of the Toronto Book Corpus, we collected the ALSS corpus as a replacement.

\subsection{Writing Relevance Training}
\label{wrtraining}
After pre-training, the model is capable of writing with correct grammar.
In the WR training stage, we adapt the model to each corpus for text continuation. The model is optimized by sum of the gradient signals from two losses (Figure \ref{fig:backprop}).
One comes from a triplet loss for distinction of \textit{coherent} continuations from negative examples, and the other is from an auxiliary token prediction loss given by the cross-entropy between predictions and the positive sentence.

\noindent The WR loss ($\mathrm{L_{WR}}$) is defined as follows:

\begin{equation}        
\label{eq:wrloss}
    \begin{aligned}
    & \mathrm{L_{WR}} =  ~\lambda \mathrm{CE}(Y^{*}, Y)  + \mathrm{TP_{cos}(a,~pos,~neg)} \\ 
    & \mathrm{TP_{cos}} (a,\text{pos},\text{neg}) = \mathrm{max}(0,1+ d^{cos}_{a,\text{pos}}-d^{cos}_{a,\text{neg}})  \\
    & \mathrm{a,~pos,~neg}=g(\mathbf{h_{X;\hat{Y}}}),~g(\mathbf{h_{X;Y}}),~g(\mathbf{h_{X;N}})
    \end{aligned}
\end{equation}

\noindent where cosine distance ($d_{cos}$) is defined as:

     \begin{equation}
        \label{cosdist}
         \begin{aligned}
         d^{cos}_{x,y} = & ~\frac{\|x\| \|y\| - x \cdot y}{2\|x\|\|y\|} 
         \end{aligned}
     \end{equation}

In Equation \ref{eq:wrloss}, $Y^*,\hat{Y}$ and $Y,N$ denote teacher-forced and greedy-decoded predictions and positive and negative continuations respectively. 
$\lambda$ is a balancing coefficient (hyperparameter) for the cross-entropy loss ($\mathrm{CE(\cdot)}$) and triplet loss ($\mathrm{TP_{cos}(\cdot)}$).
Inputs of the triplet loss, $g(\mathbf{h_{X;*}})$, are sentence representations obtained by [CLS] pooling ($\mathbf{h_{X;*}}$) as explained in the latter part of this section, mapped by $g(\cdot)$. 
If there is more than one negative example for a context, we randomly sample one among the negatives.
To sum up, WR loss is composed of an auxiliary token prediction loss and a triplet loss that regresses sentence representations.

\noindent\textbf{Decoding Strategy} We apply a simple heuristic to avoid duplication in generated sentences as in Equation \ref{eq:dup_penalty} following GLACNet~\cite{glac}.\footnote{With this heuristic, nucleus or top-k sampling~\cite{topk} show only marginal difference toward greedy decoding, so we decided to stick to greedy decoding under the heuristic.}
 
 \begin{equation}
 \label{eq:dup_penalty}
 \hat{p}(word)=p(word) \times \frac{1}{1 + k \cdot count_{word}}
 \end{equation}
 
 The value we use for $k$ in our experiments is 5. 
 Beam search is not applied since it is known to cause generic sentences in open-ended generation tasks~\cite{topp}. %\\[0.5ex]

\noindent\textbf{Sentence Representation} We use a hidden vector of special token [CLS] obtained by forward pass of $X$ and [CLS];$Y$ as a sentence representation of $X;Y$, $\mathbf{h_{X;Y}}$ (Figure \ref{fig:backprop}). 
The [CLS] pooling is mapped by $g(\cdot)$ which empirically helped triplet loss reduction. We choose the mapping amongst Hadamard product and linear projection.\footnote{We also tested other combinations such as Euclid distance metric with other $g(\cdot)$'s: identity mapping, Hadamard product, and linear projection with learnable parameters. Cosine distance metric with learnable mapping $g(\cdot)$ worked well with WR training loss.}

    % % 데이터셋 수량
    % \begin{table*}[h!]
    % \centering
    % \small
    % \begin{tabular}{|c|c|c|c|c|l|}
    % \hline
    % \textbf{Dataset}   &\textbf{Original task}   & \textbf{Use}      & \multicolumn{1}{l|}{\textbf{\begin{tabular}[c]{@{}l@{}}number of data instance \\ by splits (train/val/test)\end{tabular}}} & \multicolumn{2}{l|}{\textbf{\begin{tabular}[c]{@{}l@{}}number of negative \\ examples per instance\end{tabular}}} \\ \hline
    % \textbf{ALSS}   &   -   & pre-training   & 808.34 k / 10.11 k / -    & \multicolumn{2}{c|}{-} \\ \hline
    % \textbf{HellaSWAG}  &   Next sentence prediction  & text continuation & 39.91 k / 5.02 k / 5.02 k & \multicolumn{2}{c|}{3} \\ \hline
    % \textbf{Story Cloze Test}   &   Story completion    & text continuation & 98.16 k / 1.87 k / 1.87 k & \multicolumn{2}{c|}{$2^*$/1~(train/val~and~test)} \\ \hline
    % \end{tabular}
    % \caption{\label{table: datasets-specs} Dataset description of ALSS, HellaSWAG, and Story Cloze Test used for pre-training and the text continuation task. Note that Story Cloze Test uses \textbf{fabricated} negative examples (*) in the training split.} 
    % \end{table*}

\section{Experiments}

\subsection{Task and Datasets}
The architecture choice for our experiment is Transformer~\cite{vaswani} model.
We perform text continuation to test our approach of learning to write with \textit{coherence}. 
We use three datasets for this: i) the ALSS dataset for pre-training, which contains 4\,295 short stories (that vary from a paragraph to few pages long) collected from the public archive, ii) the Story Cloze Test, and iii) the HellaSWAG where the last two have negative examples of quality. Since the training split of the Story Cloze Test dataset has no negative examples, we randomly sample positive continuations from other stories to be used as negative examples after a sanity-check.\footnote{Random-sampled negatives only confused 3.3\% of the participants. Each participant scored as follows: 30/30, 29/30, 28/30.}

    \begin{figure}[h!]
    \includegraphics[width=\columnwidth]{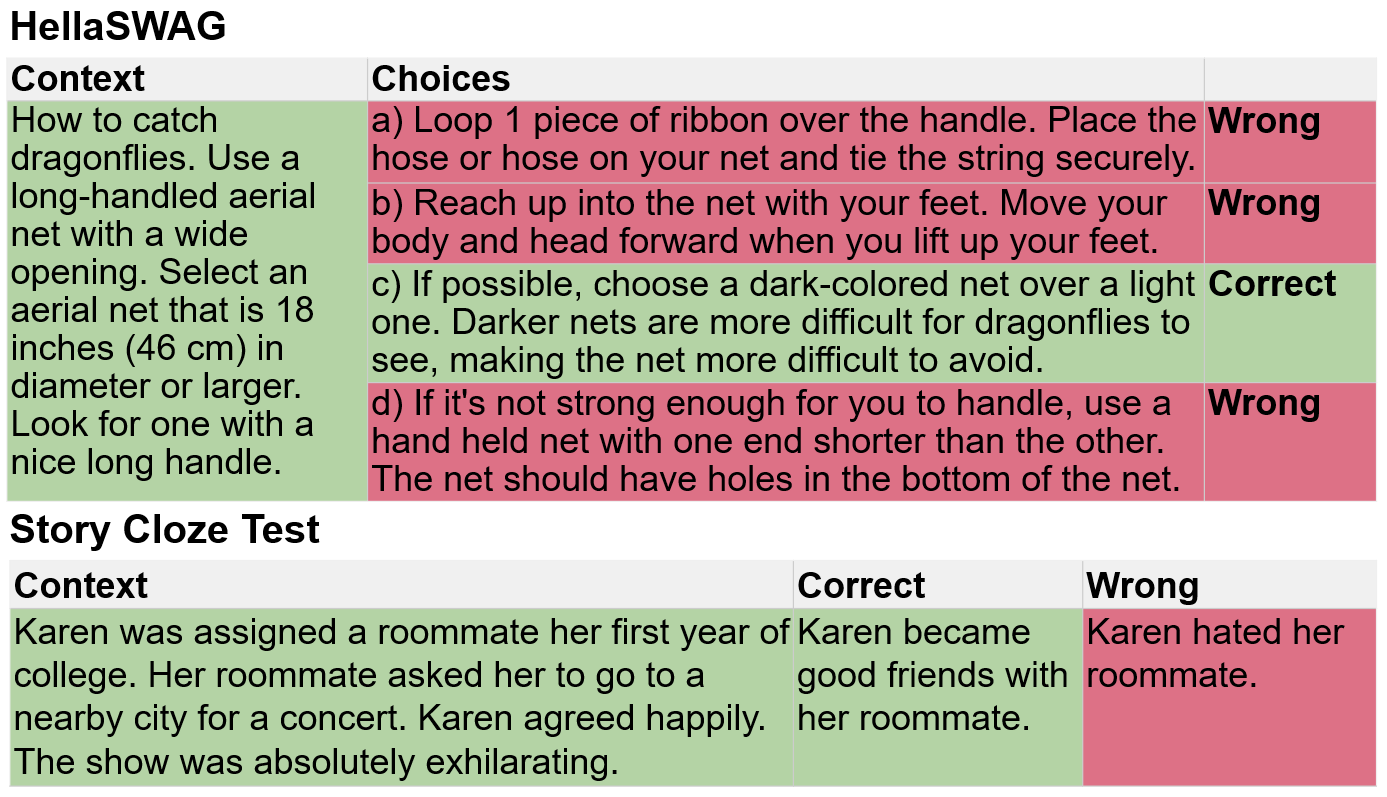}
    \caption{Examples from adaptaion corpora. Human-written negative examples are marked as ``Wrong''.}
    \label{fig:adaptation_corpora}
    \end{figure}

%Except for the ALSS corpus, which we have split into the single sentences, the other two,
HellaSWAG and Story Cloze Test datasets are composed of short pieces of writings that contain around 2-6 sentences (Figure \ref{fig:adaptation_corpora}). %Visual examples can be seen in Figure.
As a descriptive example, performing text continuation on the Story Cloze Test is often called story ending generation. %or story completion. %; given a former part of the short story, the model should generate an appropriate ending.
The Story Cloze Test dataset is originally designed as a binary choice between two human-written candidate endings where only one is correct.
Similarly, the HellaSWAG dataset is comprised of multiple-choice problems that give a context of 1-5 sentences, letting a machine choose among four candidate continuations.
The human-written negative examples of the corpora provide appropriate difficulty, thus being effective negative examples for training to learn \textit{coherence} of continuations.

%Note that the Story Cloze Test has no negative examples in the training split. 
%We used positive endings of the other stories in the split as negative examples.

\subsection{Human Evaluation}

 %describing crowd study design
 Following Stephan et al., (1981)~\cite{coherence_cohesion_writingqual}, a straightforward way of measuring \textit{coherence} is to ask readers. 
 We perform an Amazon Mechanical Turk (AMT) survey to ask people which continuation looks more natural.\footnote{We provide a preference survey rather than using Likert scale based on \cite{bestpractices}; Ranking-based assessment (including binary ranking) is more consistent.}  
 Our survey investigates sentence preference of the respondents. First, we show both context and generated continuations of two different models to participants and ask: ``What sentence do you prefer as a continuation?''. 
 Participants are also allowed to check at ``BOTH GOOD'' or ``NEITHER GOOD'' (Table \ref{table:crowdstudy_ex}).  

     % 서베이 질문 예시
    \begin{table}[h!]
    \small
    \centering
    \begin{tabular}{|l|l|}
    \hline
    Context Sentences & Choices \\
    \hline
    \multirow{8}{*}{\begin{tabular}[c]{@{}l@{}}A man breaks into a house \\ and begins to take things. \\ he takes jewelry and games, \\ some cash, and some food. \\ when the family comes home \\ they call the police. the police \\ come and investigate and \\ manage to track him.\end{tabular}} 
        & the police officer \\
        & is arrested. \\ \cline{2-2} 
        & the police come and\\
        & take the man to jail. \\ \cline{2-2}  
        &  BOTH\\ & GOOD \\ \cline{2-2} 
        &  NEITHER \\ & GOOD \\ \hline
    \end{tabular}
    \caption{\label{table:crowdstudy_ex} Question of the survey for evaluation. Annotators are asked: ``What sentence do you prefer as a continuation?''.}
    \end{table}

 In order to filter inattentive annotators out, we plant attention-check questions in the middle of each survey form which have a clearly correct choice.
 The screening rejected about 14 \% (37 out of 156) of the survey submissions.
 107 individuals participated and submitted 119 valid survey forms (average 1.11 submission per individual), each containing 27-28 questions accompanied with 3 screening questions. %(more statistics and details on the survey can be found in Appendix \ref{humaneval_stats}). %, Table \ref{sumstats} and \ref{statsbreakdown}).

     \begin{figure*}[h!] \centering
    \includegraphics[width=\linewidth ]{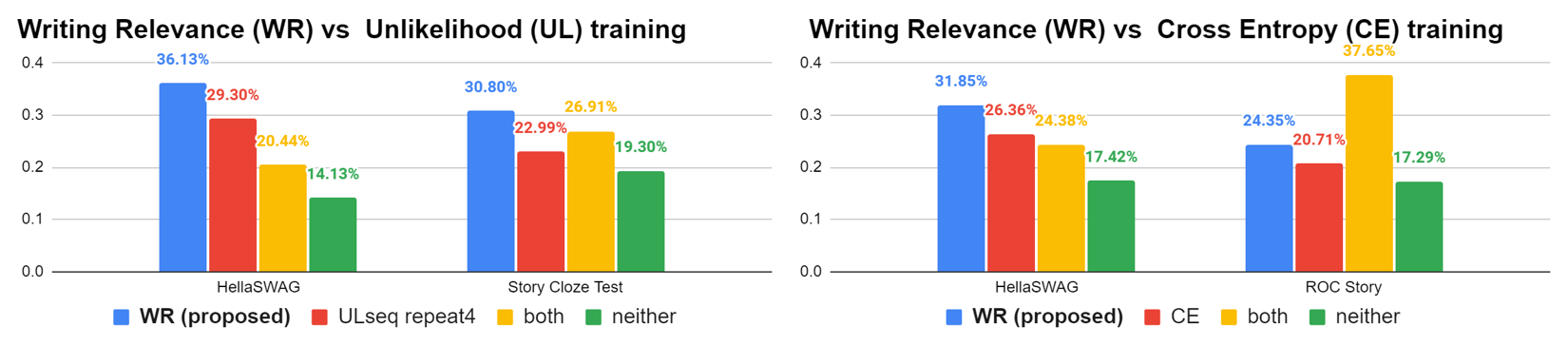}
    \caption{Preference survey results (\textit{P}=.0042, .0001, .0101, .0024). We surveyed over 107 English speaking individuals on AMT. For both comparisons (WR vs. UL and WR vs. CE), WR trained results are preferred over the corpora.}
    \label{fig:survey_chart}
    \end{figure*}

\section{Results}
We compare the performance of the NLG model that is trained with WR loss to the one that is trained using unlikelihood (UL) training loss that penalizes repetitive n-grams (n$=$4) and cross-entropy (CE) loss baselines.
Since the UL loss with repetitive n-grams as unlikely candidates outperformed taking the candidates from negative examples\footnote{Tokens that occurs only in the accompanied negative example, but not in the positive were chosen to be negative candidates.}, we take the former for the experiments.

 % Result briefing and discussions
 Figure \ref{fig:survey_chart} summarizes the preference survey conducted. 
 Continuations written by WR-trained encoder-decoder models are preferred over the other models trained with UL and CE loss on both adaptation corpora, HellaSWAG, and Story Cloze Test.
 Overall, WR $>$ UL or CE always holds.
 We conjecture that a high portion of BOTH GOOD answers in the Story Cloze Test evaluation is caused by simple sentences, and noise from fabricated negative examples.
 
 In Table \ref{table:curatedexamples}, we show continuations generated by each model.
 The continuations from WR-trained model are preferred by participants despite of lower n-gram scores. 
 While this is not always the case, for near half of the generated continuations from WR vs. UL survey, n-gram metrics failed to represent the human judgement of \textit{coherence}. This reassures that n-gram metrics are inappropriate measures for open-ended generation tasks as reported in \cite{whyweneed}.%\footnote{See Figure \ref{fig:ngramvspref} in Appendix \ref{ngramadditional} for detailed numbers. This reassures that n-gram metrics are inappropriate measures for open-ended generation tasks \cite{whyweneed}.}

\begin{table}[]
\resizebox{\columnwidth}{!}{%
\begin{tabular}{|l|l|}
\hline
\multicolumn{2}{|l|}{\begin{tabular}[c]{@{}l@{}}\textbf{Context (StoryClozeTest):} \\
a woman sits  behind a table dealing cards . she points at \\ one of the cards .\end{tabular}}                                                                                                   \\ \hline
\begin{tabular}[c]{@{}l@{}}WR (proposed) \\ (B1: 50, \underline{M: 15.6}, \textbf{50\%})\end{tabular}                                               & \begin{tabular}[c]{@{}l@{}}UL-rep.4 \\ (B1: 50, \underline{M: 18.9}, 0\%)\end{tabular}                                                         \\ \hline
\begin{tabular}[c]{@{}l@{}}she puts the cards in the \\ cards .\end{tabular}                                                & \begin{tabular}[c]{@{}l@{}}the woman is shown playing a \\ guitar and singing .\end{tabular}                                 \\ \hline\hline
\multicolumn{2}{|l|}{\begin{tabular}[c]{@{}l@{}}\textbf{Context (HellaSWAG):} \\
eric was helping his dad clear a wooded area . they were \\ 
going to put a picnic table there . all of a sudden he was \\ 
swarmed by bees . he had accidentally disturbed their nest .\end{tabular}} \\ \hline
\begin{tabular}[c]{@{}l@{}}WR (proposed) \\ (\underline{B1: 14.3}, \underline{M: 8.0}, \textbf{42.9\%})\end{tabular}                                            & \begin{tabular}[c]{@{}l@{}}UL-rep.4 \\ (\underline{B1: 18.2}, \underline{M: 14.5}, 0.0\%)\end{tabular}                                                     \\ \hline
\begin{tabular}[c]{@{}l@{}}
eric ’s dad was so upset he \\ 
had to go to the hospital. \end{tabular}                            & { \begin{tabular}[c]{@{}l@{}}
eric was able to get a bath \\
for his dad.\end{tabular}}                            \\ \hline
\end{tabular}
}
    \caption{\label{table:curatedexamples} Text continuation examples accompanied by BLEU-1 (B1), METEOR (M), and preference ratio (\%). Preference shows the \textbf{dominance} of the proposed method, while n-gram scores \underline{fail to follow} human judgements.}  
\label{tab:my-table}
\end{table}

\section{Related Work} 
There are loss-based approaches that directly incorporate negative examples into training text continuation like Large Margin LM (LMLM)~\cite{LMLM} and Unlikelihood (UL) training~\cite{UL}.
LMLM uses ranking loss to achieve better generation quality by enlarging the log-likelihood margin between the generated sentence from the negative samples.
Unlikelihood (UL) training introduces a novel unlikelihood loss function that penalizes repetitive or unlikely tokens to remedy degeneracy problems in neural generation.
Some works utilize classifiers to benefit generation. 
Holtzman et al., (2018) \cite{l2w} deploys natural language understanding (NLU) discriminators that leverage what it learned from negative examples to resolve empirical problems of neural generation. 
Gabriel et al. (2019) \cite{co-op} extends this for modeling \textit{narrative flow} in summarization task.
Similar to our approach, alignment of sentence representations for generation can also be found in Lee et al., (2021) \cite{claps} which focuses on perturbing positive and negative examples for an efficient contrastive learning of text representations for several NLG tasks.

%tackles several NLG tasks with contrastive learning loss with perturbed negative examples.

\section{Conclusion}
We proposed the writing relevance (WR) training framework, which effectively uses negative examples to improve \textit{coherence}.
WR training makes the encoder-decoder NLG model learn sentence representation in a contrastive manner to form better \textit{coherence} in its writing.
We demonstrate the potential of our WR training scheme to various domains.

% Below is an example of how to insert images. Delete the ``\vspace'' line,
% uncomment the preceding line ``\centerline...'' and replace ``imageX.ps''
% with a suitable PostScript file name.
% -------------------------------------------------------------------------
% \begin{figure}[htb]

% \begin{minipage}[b]{1.0\linewidth}
%   \centering
%   \centerline{\includegraphics[width=8.5cm]{image1}}
% %  \vspace{2.0cm}
%   \centerline{(a) Result 1}\medskip
% \end{minipage}
% %
% \begin{minipage}[b]{.48\linewidth}
%   \centering
%   \centerline{\includegraphics[width=4.0cm]{image3}}
% %  \vspace{1.5cm}
%   \centerline{(b) Results 3}\medskip
% \end{minipage}
% \hfill
% \begin{minipage}[b]{0.48\linewidth}
%   \centering
%   \centerline{\includegraphics[width=4.0cm]{image4}}
% %  \vspace{1.5cm}
%   \centerline{(c) Result 4}\medskip
% \end{minipage}
% %
% \caption{Example of placing a figure with experimental results.}
% \label{fig:res}
% %
% \end{figure}

%\vfill
%\pagebreak

\bibliographystyle{IEEEbib}
\bibliography{IEEE}

\begin{thebibliography}{10}

\bibitem{seq2seq}
Ilya Sutskever, Oriol Vinyals, and Quoc~V Le,
\newblock ``Sequence to sequence learning with neural networks,''
\newblock in {\em Advances in neural information processing systems}, 2014, pp.
  3104--3112.

\bibitem{coherence_cohesion_writingqual}
Stephen~P. Witte and Lester Faigley,
\newblock ``Coherence, cohesion, and writing quality,''
\newblock {\em College Composition and Communication}, vol. 32, no. 2, pp.
  189--204, 1981.

\bibitem{hellaswag}
Rowan Zellers, Ari Holtzman, Yonatan Bisk, Ali Farhadi, and Yejin Choi,
\newblock ``{H}ella{S}wag: Can a machine really finish your sentence?,''
\newblock in {\em Proceedings of the 57th Annual Meeting of the Association for
  Computational Linguistics}, Florence, Italy, July 2019, pp. 4791--4800,
  Association for Computational Linguistics.

\bibitem{roc}
Nasrin Mostafazadeh, Nathanael Chambers, Xiaodong He, Devi Parikh, Dhruv Batra,
  Lucy Vanderwende, Pushmeet Kohli, and James Allen,
\newblock ``A corpus and cloze evaluation for deeper understanding of
  commonsense stories,''
\newblock in {\em Proceedings of the 2016 Conference of the North {A}merican
  Chapter of the Association for Computational Linguistics: Human Language
  Technologies}, San Diego, California, June 2016, pp. 839--849, Association
  for Computational Linguistics.

\bibitem{cohesion_v_coherence}
Patricia~L. Carrell,
\newblock ``Cohesion is not coherence*,''
\newblock {\em TESOL Quarterly}, vol. 16, no. 4, pp. 479--488, 1982.

\bibitem{activitynetcaptions}
Ranjay Krishna, Kenji Hata, Frederic Ren, Li~Fei-Fei, and Juan~Carlos Niebles,
\newblock ``Dense-captioning events in videos,''
\newblock in {\em International Conference on Computer Vision (ICCV)}, 2017.

\bibitem{torontobook}
Yukun Zhu, Ryan Kiros, Rich Zemel, Ruslan Salakhutdinov, Raquel Urtasun,
  Antonio Torralba, and Sanja Fidler,
\newblock ``Aligning books and movies: Towards story-like visual explanations
  by watching movies and reading books,''
\newblock in {\em Proceedings of the IEEE international conference on computer
  vision}, 2015, pp. 19--27.

\bibitem{glac}
Taehyeong Kim, Min-Oh Heo, Seonil Son, Kyoung-Wha Park, and Byoung-Tak Zhang,
\newblock ``Glac net: Glocal attention cascading networks for multi-image cued
  story generation,''
\newblock {\em arXiv preprint arXiv:1805.10973}, 2018.

\bibitem{topk}
Angela Fan, Mike Lewis, and Yann Dauphin,
\newblock ``Hierarchical neural story generation,''
\newblock in {\em Proceedings of the 56th Annual Meeting of the Association for
  Computational Linguistics (Volume 1: Long Papers)}, Melbourne, Australia,
  July 2018, pp. 889--898, Association for Computational Linguistics.

\bibitem{topp}
Ari Holtzman, Jan Buys, Li~Du, Maxwell Forbes, and Yejin Choi,
\newblock ``The curious case of neural text degeneration,''
\newblock in {\em International Conference on Learning Representations}, 2020.

\bibitem{vaswani}
Ashish Vaswani, Noam Shazeer, Niki Parmar, Jakob Uszkoreit, Llion Jones,
  Aidan~N Gomez, {\L}ukasz Kaiser, and Illia Polosukhin,
\newblock ``Attention is all you need,''
\newblock in {\em Advances in neural information processing systems}, 2017, pp.
  5998--6008.

\bibitem{bestpractices}
Chris Van Der~Lee, Albert Gatt, Emiel Van~Miltenburg, Sander Wubben, and Emiel
  Krahmer,
\newblock ``Best practices for the human evaluation of automatically generated
  text,''
\newblock in {\em Proceedings of the 12th International Conference on Natural
  Language Generation}, 2019, pp. 355--368.

\bibitem{whyweneed}
Jekaterina Novikova, Ond{\v{r}}ej Du{\v{s}}ek, Amanda Cercas~Curry, and Verena
  Rieser,
\newblock ``Why we need new evaluation metrics for {NLG},''
\newblock in {\em Proceedings of the 2017 Conference on Empirical Methods in
  Natural Language Processing}, Copenhagen, Denmark, Sept. 2017, pp.
  2241--2252, Association for Computational Linguistics.

\bibitem{LMLM}
Jiaji Huang, Yi~Li, Wei Ping, and Liang Huang,
\newblock ``Large margin neural language model,''
\newblock in {\em Proceedings of the 2018 Conference on Empirical Methods in
  Natural Language Processing}, Brussels, Belgium, Oct.-Nov. 2018, pp.
  1183--1191, Association for Computational Linguistics.

\bibitem{UL}
Sean Welleck, Ilia Kulikov, Stephen Roller, Emily Dinan, Kyunghyun Cho, and
  Jason Weston,
\newblock ``Neural text generation with unlikelihood training,''
\newblock in {\em International Conference on Learning Representations}, 2020.

\bibitem{l2w}
Ari Holtzman, Jan Buys, Maxwell Forbes, Antoine Bosselut, David Golub, and
  Yejin Choi,
\newblock ``Learning to write with cooperative discriminators,''
\newblock in {\em Proceedings of the 56th Annual Meeting of the Association for
  Computational Linguistics (Volume 1: Long Papers)}, Melbourne, Australia,
  July 2018, pp. 1638--1649, Association for Computational Linguistics.

\bibitem{co-op}
Saadia Gabriel, Antoine Bosselut, Ari Holtzman, Kyle Lo, Asli
  {\c{C}}elikyilmaz, and Yejin Choi,
\newblock ``Cooperative generator-discriminator networks for abstractive
  summarization with narrative flow,''
\newblock {\em CoRR}, vol. abs/1907.01272, 2019.

\bibitem{claps}
Seanie Lee, Dong~Bok Lee, and Sung~Ju Hwang,
\newblock ``Contrastive learning with adversarial perturbations for conditional
  text generation,''
\newblock in {\em International Conference on Learning Representations}, 2021.

\end{thebibliography}

\end{document}